\documentclass[letterpaper]{article} 
\usepackage[preprint]{aaai23}  
\usepackage{times}  
\usepackage{helvet}  
\usepackage{courier}  
\usepackage[hyphens]{url}  
\usepackage{graphicx} 
\usepackage{natbib}  
\usepackage{caption} 
\usepackage{algorithm}
\usepackage{algorithmic}
\usepackage{amsmath, amsthm, amssymb}
\usepackage{cleveref}
\usepackage{booktabs}
\usepackage{xspace}
\usepackage{makecell}
\usepackage{bm}
\usepackage{cleveref}
\usepackage{multirow}
\usepackage{subfigure}
\usepackage{amssymb}

\newtheorem{theorem}{Theorem}
\newtheorem{lemma}{Lemma}

\usepackage{newfloat}
\usepackage{listings}
\DeclareCaptionStyle{ruled}{labelfont=normalfont,labelsep=colon,strut=off} 
\floatstyle{ruled}
\newfloat{listing}{tb}{lst}{}
\floatname{listing}{Listing}
\title{Disentangled Generation with Information Bottleneck for Few-Shot Learning}
\author{
    Zhuohang Dang, Jihong Wang, Minnan Luo, Chengyou Jia, Caixia Yan, Qinghua Zheng
}
\affiliations{
dangzhuohang@stu.xjtu.edu.cn
}

\usepackage{bibentry}

\makeatletter
\DeclareRobustCommand\onedot{\futurelet\@let@token\@onedot}
\def\@onedot{\ifx\@let@token.\else.\null\fi\xspace}
\def\eg{\emph{e.g}\onedot} 
\def\ie{\emph{i.e}\onedot}

\def\etc{\emph{etc}\onedot}

\makeatother

\begin{document}

\maketitle

\begin{abstract}
Few-shot learning (FSL), which aims to classify unseen classes with few samples, is challenging due to data scarcity. Although various generative methods have been explored for FSL, the entangled generation process of these methods exacerbates the distribution shift in FSL, thus greatly limiting the quality of generated samples.
To these challenges, we propose a novel Information Bottleneck (IB) based Disentangled Generation Framework for FSL, termed as DisGenIB, that can simultaneously guarantee the discrimination and diversity of generated samples. Specifically, we formulate a novel framework with information bottleneck that applies for both disentangled representation learning and sample generation. Different from existing IB-based methods that can hardly exploit priors, we demonstrate our DisGenIB can effectively utilize priors to further facilitate disentanglement.
We further prove in theory that some previous generative and disentanglement methods are special cases of our DisGenIB, which demonstrates the generality of the proposed DisGenIB. Extensive experiments on challenging FSL benchmarks confirm the effectiveness and superiority of DisGenIB, together with the validity of our theoretical analyses. Our codes will be open-source upon acceptance.
\end{abstract}

\section{Introduction}
Few-shot Learning (FSL) that recognizes the unseen classes with few labeled samples is challenging, since models trained with very few samples notoriously lead to underfitting or overfitting problems \cite{liu2020prototype}.
There are mainly two routes to address the FSL problem: the algorithm-based methods and the data augmentation-based methods. For the algorithm-based methods  \cite{sun2019meta,liu2021learning}, researchers try to learn robust task-agnostic feature representations or an effective meta-learner that can quickly adapt models' parameters with few samples. Nonetheless, these methods severely suffer from the intra-class variance of labeled samples, due to the data scarcity of FSL. 
In this paper, we set our sights on the data augmentation-based methods, which tackle data scarcity in a straightforward way. These methods typically augment data by training generative models on seen classes for fitting data distribution  \cite{li2020adversarial,luo2021few}, and then utilize the learned knowledge to generate samples on the unseen classes. Some works also introduce external knowledge, \eg, semantics or attributes, as label-related priors to enhance the generation \cite{xu2022generating}. 

\setlength{\belowcaptionskip}{-0.2cm}
\begin{figure}[!t]
    \centering
    \subfigure[Entangled generation methods.]{
    \begin{minipage}[t]{0.99\linewidth}
    	\begin{center}
    		\includegraphics[width=1\linewidth]{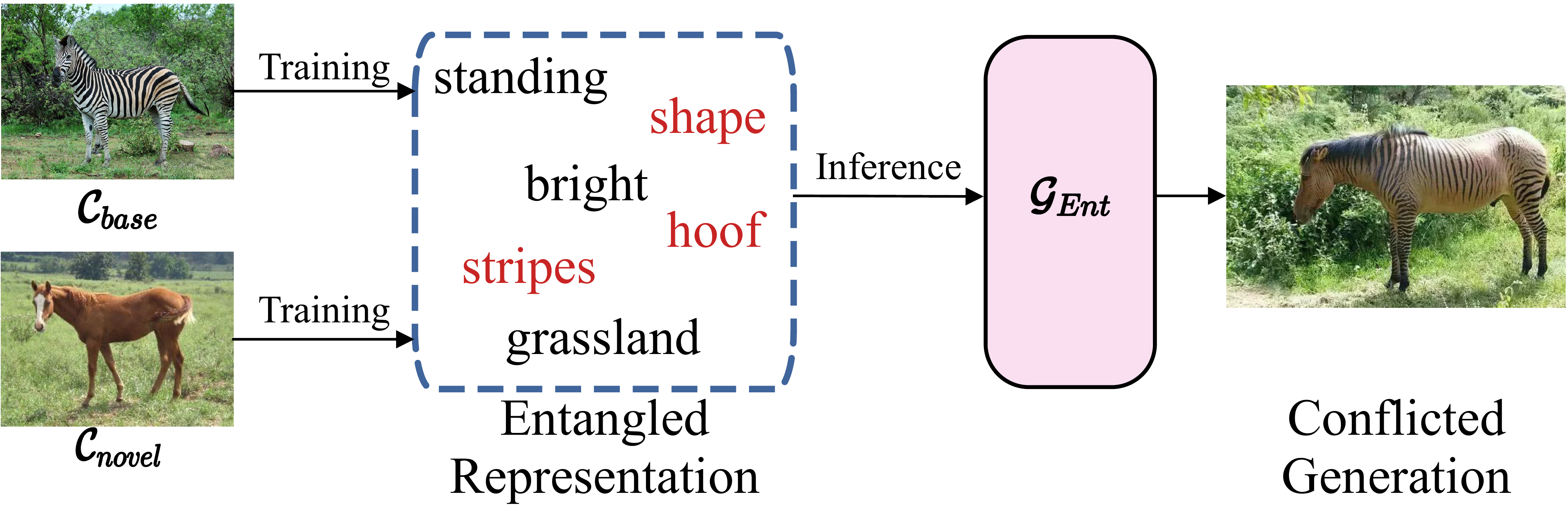}
    	\end{center}
    	\label{fig: entangled}
    \end{minipage}%
    }%
    \\
    \subfigure[The proposed DisGenIB.]{
    \begin{minipage}[t]{0.99\linewidth}
    	\begin{center}
    		\includegraphics[width=1\linewidth]{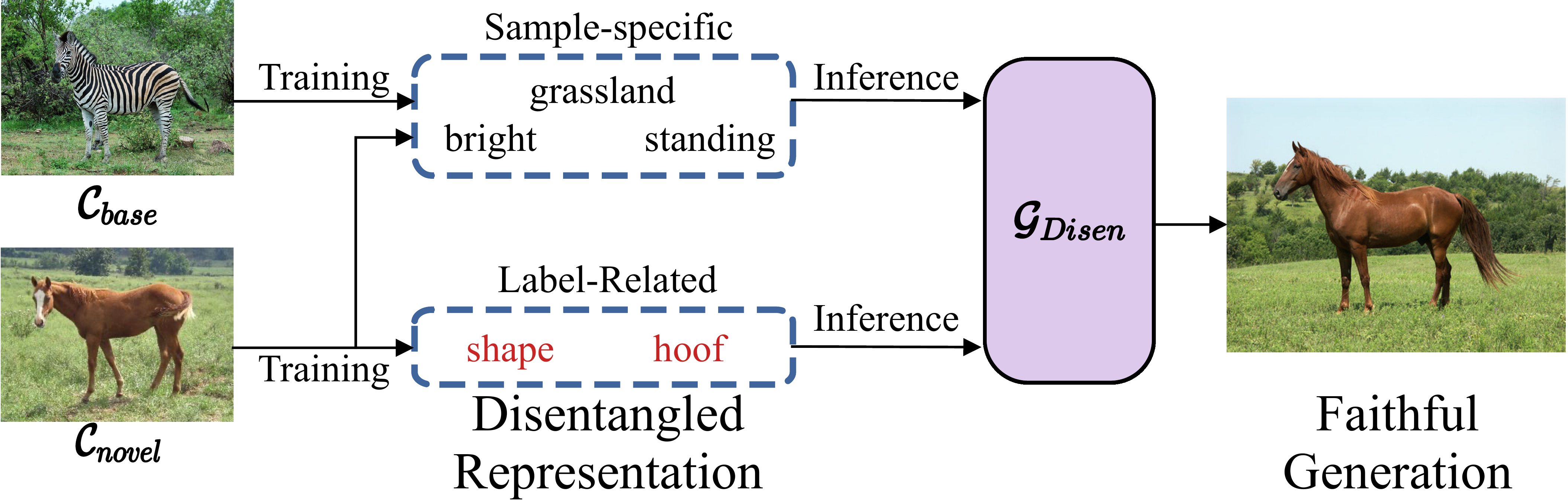}
    	\end{center}
    	\label{fig: disentangled}
    \end{minipage}%
    }%
\caption{Illustration of unseen sample generation in FSL, where $\mathcal{C}_{base}$ and $\mathcal{C}_{novel}$ denote classes of seen and unseen; $\mathcal{G}_{Ent}$ and $\mathcal{G}_{Disen}$ indicate the generator of entangled and disentangled. The red words refer to label-related information, while the black ones denote sample-specific information.}
\end{figure}
\setlength{\belowcaptionskip}{0cm}

Notably, previous data augmentation-based methods usually fail to disentangle two complementary components: label-related and sample-specific information, which are roughly decomposed from image information. This problem exacerbates the distribution shift between seen and unseen categories in FSL, and thus misleads the generated samples into deviating from true data distribution. Here, the label-related information corresponds to a specific class, inheriting discriminative information for classification. The sample-specific information that characterizes the diversity of samples, \eg, view, light and pose, is independent of labels and generalizable to all classes. 
As shown in \Cref{fig: entangled}, the entangled generation process misuses some label-related information (\ie, stripes of zebra)  to generate unseen samples (\ie, horse), which leads to conflicted generations. 
Some methods \cite{xu2021variational,cheng2021disentangled} regard features for classification as label-related information to disentangle features. However, according to \cite{yue2020interventional}, it is infeasible since feature extractors and classifiers fail to obtain comprehensive and accurate label-related features.
Thus, it is significantly necessary to develop an effective disentangled generation framework for FSL.

Recently, various methods  \cite{jeon2021ib,pan2021disentangled} have been proposed for disentangled representation learning with Information Bottleneck (IB)  \cite{tishby2000information} that can extract label-related information from input.
For example,  \cite{pan2021disentangled} employs IB as training objective for the encoder-decoder architecture, which encodes the class-specific information into label-related features for compression, while eliminating discriminative information from sample-specific features for disentanglement.
Some downstream tasks, \eg, classification, detection and generation, are followed to evaluate the efficacy of the disentangled representations  \cite{bao2021disentangled,kim2021distilling,uddin2022federated}.
However, these methods rely on a two-stage process for generation,
which fails to jointly optimize the representation learning and sample generation for optimality.
Moreover, these methods are difficult to utilize the widely used priors in FSL to facilitate the generation of unseen classes.

In this paper, we propose a unified framework for disentangled representation learning and sample generation based on IB, termed as DisGenIB, which can also effectively utilize priors to facilitate disentanglement. As shown in \Cref{fig: disentangled}, our DisGenIB can generate unseen samples with label-related information from the labeled sample and diverse sample-specific information from various samples. The former ensures the discrimination of specific class, while the latter enriches the diversity of generated samples.
Since the mutual information involved in IB cannot be optimized directly, we formulate a tight and tractable bound via variational inference. We also prove in theory that disentangled IB (DisenIB) and conditional variational autoencoder (CVAE) are special cases of the proposed DisGenIB, which demonstrate the superiority of DisGenIB on generality.
In summary, the contributions of this paper are as follows:
\begin{itemize}
    \item We propose an IB-based disentangled generation framework to synthesize samples of unseen classes for FSL. 
    It is capable of simultaneously maintaining the discrimination and diversity of generated samples from principle. To the best of our knowledge, this is the first work that explores IB for disentangled generation for FSL.

    \item In theory, we derive a tight and tractable bound to solve the optimization problem of DisGenIB, and prove that some previous disentanglement and generative methods, including DisenIB, CVAE and AVAE, are special cases of the proposed DisGenIB.
    
    \item Extensive experiments demonstrate the significant performance improvement of our DisGenIB, which outperforms the state-of-the-arts by more than 5\% and 1\% in 5-way 1-shot and 5-shot FSL tasks, respectively. 
  \end{itemize}

\section{Related Works}
\subsection{Few-Shot Learning}
FSL aims to learn concepts of unseen classes with very few labeled samples, mainly in the following two ways. (1) Algorithm-based methods that learn a feasible feature space or a meta-learner that can quickly adapt to new tasks with few samples. 
For example, to learn an effective meta-learner, MAML \cite{finn2017model} effectively learns models with good potential by maximizing the parameters' sensitivity to the loss of new tasks. 
To rectify the decision boundaries, EPNet \cite{rodriguez2020embedding} utilizes embedding propagation to smooth the manifold. 
While for the feasible feature space, Prototypical Networks  \cite{snell2017prototypical} alleviates the intra-class bias by viewing the feature centroid as class prototype, where new samples can be classified via nearest neighbors. BD-CSPN \cite{liu2020prototype} rectifies the prototype by assigning pseudo-labels to samples with high confidence. (2) Data augmentation-based methods try to directly generate unseen samples for FSL. AFHN \cite{li2020adversarial} generates samples with generative adversarial networks pre-trained on seen classes. 
There are some methods introducing priors to further improve quality of generated samples. To name a few, R-SVAE  \cite{xu2022generating} uses class semantics to formalize a conditional generator with attributes. MM-Res \cite{pahde2021multimodal} explores multi-modal consistency between images and texts to enrich the diversity of generated samples. 

However, with prior or not, these methods fail to achieve the disentanglement during generation process, thus suffering from the distribution shift.
In contrast, our DisGenIB provides a unified framework from the information theoretic perspective. Not only can it ensure the disentangled representation learning and sample generation, but also effectively leverages priors, thus simultaneously maintaining the discrimination and diversity of generated samples.

\subsection{Disentangled Representation Learning}
Following the encoder-decoder framework, disentangled representation learning aims to learn mutually independent representations from input. For example, $\beta$-VAE  \cite{higgins2016beta} learns factorized representations by adjusting the hyperparameter $\beta$, which controls the trade-off between reconstruction performance and disentanglement. CausalVAE  \cite{yang2021causalvae} employs structured causal model to recover the independent latent factors with causal structure. InfoGAN  \cite{chen2016infogan} learns disentangled representation through maximizing the mutual information between latent factors and observations.
Moreover, IB is introduced to disentanglement methods as a regularizer to refine the learned representations  \cite{gao2021information,jeon2021ib}. However, these methods fail to demonstrate the power of disentanglement in principle. In contrast, DisGenIB theoretically analyzes the disentanglement performance, together with proving the conditional variational autoencoder (CVAE) \cite{sohn2015learning} is special case of DisGenIB.

\section{Disentangled Generative Information Bottleneck}
Given input sample $X$ with its label $Y$, let $(A,Z)$ be a pair of disentangled representations of $X$, where $A$ is the representation of label-related information inherited from $X$, while $Z$ encodes the sample-specific information. Variables $A$ and $Z$ are complementary to each other.
From an information-theoretic perspective, we propose DisGenIB for disentangled representation learning and sample generation in a unified framework by solving optimization problem
\begin{align}\label{eq:DisGenIB}
        \max_{A,Z}
        \mathcal{L}_{DisGenIB}= &\underbrace{I(X;A,Z)}_{Reconstruction}+\underbrace{I(A;Y)-\beta I(X;A)}_{Compression}\notag\\
        &-\underbrace{(1+\alpha)I(Y;Z)}_{Disentanglement},
\end{align}
where maximizing mutual information $I(X;A,Z)$ encourages disentangled representation pair $(A,Z)$ to be sufficient for input $X$.
Maximizing the second term $I(A;Y)-\beta I(X;A)$ follows traditional IB to find a maximally compressed $A$ of $X$, which preserves as much as possible the information on $Y$  \cite{tishby2000information}.
For the third term, minimizing mutual information $I(Y;Z)$ aims to eliminate the label-related information from sample-specific representation $Z$, so as to acquire disentanglement between variables $A$ and $Z$.
Hyperparameter $\alpha$ and $\beta$ control the degree of disentanglement and compression, respectively.

\subsection{Estimation of DisGenIB}
Note that mutual information based objective ($L_{DisGenIB}$) is intractable to directly compute and optimize since mutual information usually consists of integral on high-dimensional data. 
To this issue, we follow previous work  \cite{alemi2016deep}, and leverage variational inference to estimate corresponding bounds for three terms. 
Due to space limitations, the detailed proof is shown in supplementary materials.

\paragraph{Reconstruction term.}
Let $Q_{\phi_1}(X|A,Z)$ be a variational approximation to true posterior $P(X|A,Z)$ with parameter $\phi_1$, we formulate a tractable lower bound for $I(X;A,Z)$ as
{\small{
    \begin{align}\label{eq:I(X;ZA)}
    &I(X;A,Z) =\int P(X,A,Z)\log\frac{P(X|A,Z)}{P(X)} \,da\,dx\,dz\\
    &\geq \int P(X,A,Z)\log Q_{\phi_1}(X|A,Z) \,da\,dx\,dz+H(X),\notag
    \end{align}}}
where $H(X)$ is constant that can be ignored.

\paragraph{Compression term.}
Let $Q_\gamma(Y|A)$ be a variational approximation to true posterior $P(Y|A)$ with parameter $\gamma$, we derive a tractable lower bound of $I(A;Y)$ as
\begin{equation*}
    \begin{aligned}
        I(A;Y) &= \int P(A,Y)\log \frac{P(Y|A)}{P(Y)} \,da\,dy \\
        &\geq\int P(A,Y)\log Q_\gamma(Y|A) \,da\,dy + H(Y),
    \end{aligned}
\end{equation*}
where $H(Y)$ is a constant that can be ignored.

Following the strategy used in  \cite{wang2021learning}, we use Contrastive Log-ratio Upper Bound (CLUB)  \cite{cheng2020club} as an estimator of mutual information $I(X;A)$, \ie,
{\small{\begin{equation*}
\begin{aligned}
I(X;A)\leq
\mathrm{I}_{\mathrm{vCLUB}}(X;A)
&= \mathbb{E}_{P(X, A)}[\log Q_{\theta_A}(A|X)] \\&
\ \ \ \ \ \ -\mathbb{E}_{P(X)} \mathbb{E}_{P(A)}[\log Q_{\theta_A}(A |X)],
\end{aligned}
\end{equation*}}}where variational distribution $Q_{\theta_A}(A|X)$ is an approximation to true posterior $P(A|X)$ with parameter $\theta_A$. 
\begin{figure}
    \centering
    \includegraphics[width=1\linewidth]{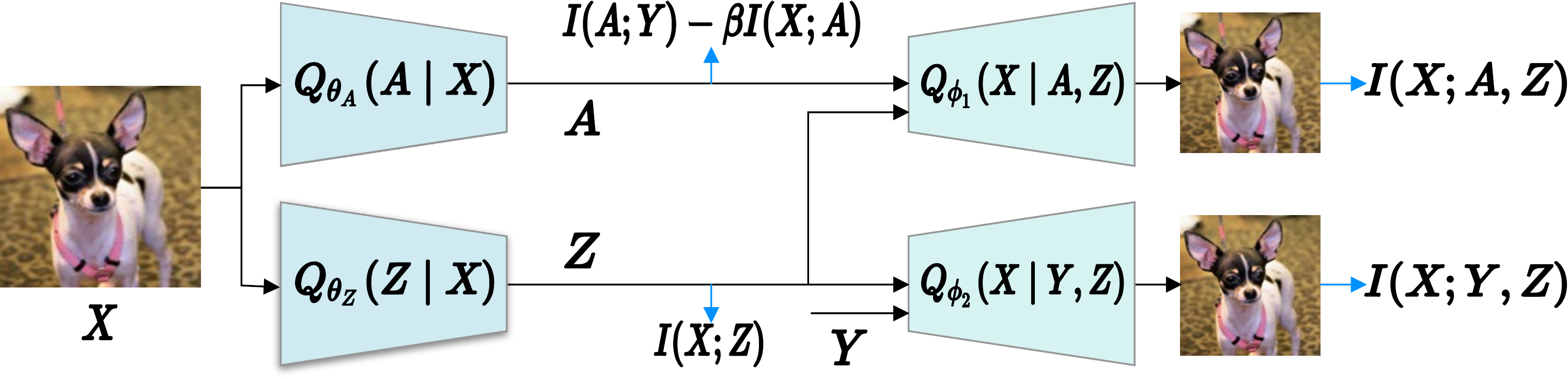}
    \caption{An overview training procedure of DisGenIB, where blue arrows denote corresponding constraints.}
    \label{fig:overview}
\end{figure}

\paragraph{Disentanglement term.}
It is difficult to estimate an upper bound of mutual information $I(Y;Z)$ since the sample-specific representation $Z$ is independent of label variable $Y$. To this end, we derive in \Cref{prop:chain rule} an equivalent training objective based on Markov chain $Y\leftrightarrow X \leftrightarrow Z$.
\begin{theorem}
    \label{prop:chain rule}
        Let $X,Y,Z$ be random variables with joint distribution $P(X,Y,Z)$. Assuming $P(X,Y,Z)$ follows Markov chain $Y\leftrightarrow X\leftrightarrow Z$, \ie, $P(X,Y,Z)=P(X)P(Y|X)P(Z|X)$, we have
       \begin{align}
           I(Y;Z) = I(X;Z) + I(X;Y) - I(X;Y,Z),
       \end{align}
       where $I(X;Y)$ is a constant. 
    \begin{proof}
        \renewcommand{\qedsymbol}{}
        Please see supplementary material for proof.
    \end{proof}
\end{theorem}
Similar to the estimation of mutual information $I(X;A)$ and $I(X;A,Z)$ above, the corresponding bounds of $I(X;Z)$ and $I(X;Y,Z)$ can be derived with the help of varitional distributions $Q_{\theta_Z}(Z|X)$ and $Q_{\phi_2}(X|Y,Z)$, respectively (see supplementary for detailed proof and derivation).

In summary, we illustrate the overview training procedure of DisGenIB in \Cref{fig:overview}. In detail, $Q_{\theta_Z}(Z|X)$ and $Q_{\theta_A}(A|X)$ parameterized with $\theta_Z$ and $\theta_A$ can be viewed as two encoders to infer the sample-specific representation $Z$ and label-related representation $A$ from Gaussian distribution of input $X$. 
$Q_{\phi_1}(X|A,Z)$ and $Q_{\phi_2}(X|Y,Z)$ parameterized with $\phi_1$ and $\phi_2$ can be viewed as two decoders to reconstruct input $X$ from the joint distribution $P(A,Z)$ and $P(Y,Z)$, respectively. The last term $Q_{\gamma}(Y|A)$ that is parameterized by $\gamma$, can be regarded as a classifier to predict label $Y$ according to the label-related representation $A$.

\subsection{Involving priors into DisGenIB}
Note that the objective of DisGenIB in \Cref{eq:DisGenIB} is formulated without any priors.
In point of fact, there are some available priors of label-related information, such as class semantics, hierarchy and attributes \cite{li2020boosting,xu2022generating}. 
These priors provide critical information of unseen classes to guide the training process of FSL, thereby significantly improving performance. 
It is therefore reasonable to assume that the label-related information follows priors, which are distributed as $P(A)$.
\footnote{Note that we don't explore the case where the prior distribution $P(Z)$ is provided, since $Z$ only contains the sample-specific information that is less operational meaning.}
In this sense, we utilize prior distribution $P(A)$ and derive the following \Cref{prop:inequality} and \Cref{prop:DisGenIB_prior} for an efficient reduction of the proposed DisGenIB.

\begin{lemma}
    \label{prop:inequality}
        Let $Y$ be label variable, $A$ be variable containing label-related information and $Z$ be variable with sample-specific information.
        Given prior distribution $P(A)$ of label-related information $A$, the following inequality holds
        \begin{align}
             I(A;Z) \geq I(Y;Z).
        \end{align}
    \begin{proof}
        \renewcommand{\qedsymbol}{}
        Please see supplementary for detailed proof.
    \end{proof}
\end{lemma}

\begin{theorem}
    \label{prop:DisGenIB_prior}
        Given prior distribution $P(A)$ and Markov chain $A\leftrightarrow X\leftrightarrow Z$, the optimization problem (\ref{eq:DisGenIB}) is reduced to optimization problem 
    \begin{equation}\label{eq:DisGenIB_prior}
        \max_{Z}\mathcal{L}_{DisGenIB}^{Prior} = I(X;A,Z)-\alpha^\prime I(X;Z),
\end{equation}
where hyperparameter $\alpha^\prime = \frac{1+\alpha}{2+\alpha}$ controls the trade-off between reconstruction and disentanglement.
\begin{proof}
\renewcommand{\qedsymbol}{}
With prior distribution $P(A)$, the compression term in \Cref{eq:DisGenIB} turns to be constant $C$ since the values of mutual information $I(X;A)$ and $I(Y;A)$ are determined.

Moreover, according to \Cref{prop:inequality}, we have
{\small{
\begin{align}
    \mathcal{L}_{DisGenIB}
    \geq& I(X;A,Z)-(1+\alpha)I(A;Z)+C\notag \\
    =& I(X;A,Z)+C\notag \\
    -&(1+\alpha)(I(X;A)+I(X;Z)-I(X;A,Z))\\
    =& (2+\alpha)I(X;A,Z)-(1+\alpha)I(X;Z)+C^\prime,
\end{align}}}
where $C^\prime=C-(1+\alpha)I(X;A)$ is a constant that can be ignored. The derivation of $I(A;Z)$ in Equation (6) holds due to the \Cref{prop:chain rule} and Markov chain $A\leftrightarrow X\leftrightarrow Z$ \cite{kim2018disentangling}.
With the formulations of $\alpha^\prime$ and $\mathcal{L}_{DisGenIB}^{Prior}$, the proof is completed. 
\end{proof}
\end{theorem}
Note that the terms in $\mathcal{L}_{DisGenIB}^{Prior}$ have been estimated using \Cref{eq:I(X;ZA)} and vCLUB, respectively. 
Therefore, given prior distribution $P(A)$, the proposed DisGenIB degenerates to an encoder $I(X;Z)$ and a decoder $I(X;A,Z)$, which is evidently more efficient than the case without prior.

\begin{table*}[!t]
\tabcolsep = 4 pt
    \begin{tabular}{lcccccc}
    \toprule 
    \multicolumn{1}{l}{\multirow{2}{*}{\textbf{Method}}}
    & \multicolumn{1}{c}{\multirow{2}{*}{\makecell{\textbf{Using} \\ \textbf{Priors}}}} 
    & \multicolumn{1}{c}{\multirow{2}{*}{\textbf{Backbone}}} 
    & \multicolumn{2}{c}{\textbf{miniImageNet}} 
    & \multicolumn{2}{c}{\textbf{tieredImageNet}} 
    \\ 
    \cmidrule(l){4-7}
    \multicolumn{1}{c}{}
    & \multicolumn{1}{c}{}         
    & \multicolumn{1}{c}{}   
    & \textbf{$5$-way $1$-shot} & \textbf{$5$-way $5$-shot}  & \textbf{$5$-way $1$-shot}  & \textbf{$5$-way $5$-shot} \\  
    \midrule
    \midrule
    CGCS  \cite{gao2021curvature}                   & No                                 & ResNet-12                           & $67.02 \pm 0.20\%$               & $82.32 \pm 0.14\%$             & $71.66 \pm 0.23\%$                 & $75.50 \pm 0.15\%$                 \\
    RENet  \cite{kang2021relational}                   & No                                 & ResNet-12                           & $67.60 \pm 0.44\%$               & $82.58 \pm 0.30\%$             & $71.61 \pm 0.51\%$                 & $85.28 \pm 0.35\%$                 \\
    RFS  \cite{tian2020rethinking}                     & No                                 & ResNet-12                          & $64.82 \pm 0.60\%$              & $82.14 \pm 0.43\%$            & $71.52 \pm 0.69\%$                 & $86.03 \pm 0.49\%$                 \\
    InvEq  \cite{rizve2021exploring}                   & No                                 & ResNet-12                           & $67.28 \pm 0.80\%$              & $84.78 \pm 0.52\%$             & $71.87 \pm 0.89\%$                 & $86.82 \pm 0.58\%$                 \\
    DeepEMD  \cite{zhang2020deepemd}                 & No                                 & ResNet-12                          & $65.91 \pm 0.82\%$              & $82.41 \pm 0.56\%$             & $71.16 \pm 0.87\%$                 & $86.03 \pm 0.58\%$                 \\
    AFHN  \cite{li2020adversarial}                    & No                                 & ResNet-18                          & $62.38 \pm 0.72\%$              &  $78.16 \pm 0.56\%$            &       -           &      -            \\
    infoPatch  \cite{liu2021learning}               & No                                 & ResNet-12                          & $67.67 \pm 0.45\%$              & $82.44 \pm 0.31\%$             & $71.51 \pm 0.52\%$                 & $85.44 \pm 0.35\%$                 \\
    FRN  \cite{wertheimer2021few}                     & No                            & ResNet-12                     & $66.45 \pm 0.19\%$              &  $82.83 \pm 0.13\%$            & $72.06 \pm 0.22\%$                 & $86.89\pm0.14\%$                 \\
    PAL  \cite{ma2021partner}                    & No                                 & ResNet-12                          & $69.37 \pm 0.64\%$              & $84.40 \pm 0.44\%$             & $72.25 \pm 0.72\%$                 & $86.95 \pm 0.47\%$                 \\
    ODE  \cite{xu2021learning}                    & No                                 & ResNet-12                          & $67.76 \pm 0.46\%$              & $82.71 \pm 0.31\%$             & $71.89 \pm 0.52\%$                 & $85.96 \pm 0.35\%$                 \\
    MeTAL  \cite{baik2021meta}                    & No                                 & ResNet-12                          & $59.64 \pm 0.38\%$              & $76.20 \pm 0.19\%$             & $63.89 \pm 0.43\%$                 & $80.14 \pm 0.40\%$                 \\
    HGNN  \cite{yu2022hybrid}                    & No                                 & ResNet-12                          & $67.02 \pm 0.20\%$              & $83.00 \pm 0.13\%$             & $72.05 \pm 0.23\%$                 & $86.49 \pm 0.15\%$                 \\
    \midrule
    DisGenIB                    & No                                 & ResNet-12                          & \textbf{76.31} $\pm$ \textbf{0.84}\%              & \textbf{85.56} $\pm$ \textbf{0.53}\%             & \textbf{76.92} $\pm$ \textbf{0.90}\%                 & \textbf{87.32} $\pm$ \textbf{0.59}\%                \\
    \midrule
    \midrule
    TriNet  \cite{chen2019multi}                  & Yes                                 & ResNet-18                          & $58.12 \pm 1.37\%$              & $76.92 \pm 0.69\%$             &        -          &       -           \\
    TRAML  \cite{li2020boosting}               & Yes                                 & ResNet-12                          & $67.10 \pm 0.52\%$              & $79.54 \pm 0.60\%$             &    -             &       -         \\
    TADAM  \cite{xing2019adaptive}                & Yes                                 & ResNet-12                          & $65.21 \pm 0.30\%$              & $75.20 \pm 0.27\%$             & $69.08 \pm 0.47\%$                 & $82.58 \pm 0.31\%$                 \\
    FSLKT  \cite{peng2019few}               & Yes                                 &  ConvNet-128                         & $64.42 \pm 0.72\%$              & $74.16 \pm 0.56\%$             &        -          &   -          \\
    MetaDT  \cite{zhang2022metadt}                  & Yes                                  & ResNet-12                          &  $69.08 \pm 0.73\%$             & $83.40 \pm 0.51\%$             & $70.56 \pm 0.90\%$                 & $85.17 \pm 0.56\%$                 \\
    SVAE  \cite{xu2022generating}                  & Yes                                  & ResNet-12                          &  $73.01 \pm 0.24\%$             & $83.13 \pm 0.40\%$             & $76.36 \pm 0.65\%$                 & $85.65 \pm 0.50\%$                 \\
    \midrule
    DisGenIB + Prior                    & Yes                                 & ResNet-12                          & \textbf{79.56} $\pm$ \textbf{0.81}\%              & \textbf{86.18} $\pm$ \textbf{0.52}\%             & \textbf{77.60} $\pm$ \textbf{0.89}\%                 & \textbf{87.38} $\pm$ \textbf{0.58}\%\\
    \bottomrule
    \end{tabular}
    \caption{Comparison with SOTA methods on miniImageNet and tieredImageNet.}
    \label{tab:mini_tiered_results}
    \end{table*}

\section{Theoretical Analysis}
In this section, we demonstrate the generality of DisGenIB in theory that previous disentanglement  \cite{pan2021disentangled} and generative methods  \cite{sohn2015learning,xu2021variational} are special cases of our DisGenIB.

\subsection{Connection with Disentangled IB}
Disentangled IB (DisenIB)  \cite{pan2021disentangled} develops IB principle for optimal disentangled representation learning, by solving the following optimization problem
\begin{equation}
    \min_{A,Z} \mathcal{L}_{DisenIB} = -I(A;Y)-I(X;Z,Y)+I(A;Z).
\end{equation}
Compared to previous disentanglement methods  \cite{nie2020semi,jeon2021ib}, DisenIB is an effective and interpretable model due to its consistency on maximum compression  \cite{pan2021disentangled}.
We prove in \Cref{th:equivalent} that DisenIB is a special case of our DisGenIB, and thus demonstrate the effectiveness of DisGenIB on disentanglement (see supplementary for proof in detail).
\begin{theorem}\label{th:equivalent}
	For disentangled representation learning, DisGenIB is reduced to DisenIB when $\alpha=0$ and $\beta=1$.
\end{theorem}

In spite of the similarity, our DisGenIB further overcomes some defects of DisenIB. Note that DisenIB tackles the disentanglement term $I(A;Z)$ via adversarial training with complicated data sampling and rearrangement; Moreover, it is significantly difficult to achieve Nash equilibrium in adversarial training. In contrast, our disentanglement strategy is more effective and stable because the compression and disentanglement terms are optimized directly via reparameterization trick with standard sample batches.

\subsection{Connection with Conditional Varational Autoencoder (CVAE)}
CVAE  \cite{sohn2015learning,verma2018generalized} is a dominant generative model, which is usually used in FSL to synthesize extra samples of unseen classes based on knowledge of seen classes, such as SVAE \cite{xu2022generating}, DCVAE \cite{zhang2021dizygotic} and VFD \cite{xu2021variational}, \etc. 
It is formulated as solving the following optimization problem
\begin{equation}\label{eq:cvae loss}
    \begin{aligned}
    \max_{Z}\mathcal{L}_{CVAE}=& 
    \mathbb{E}_{P(X,Y)}\mathbb{E}_{P(Z| X)}\left[\log Q(X|Z, Y)\right]
    \\&
    -\beta
    \mathbb{E}_{P(X)}\left[D_{KL}(P(Z|X)\|H(Z))\right],
    \end{aligned}
    \end{equation}
where $D_{KL}$ refers to KL-divergence; $H(Z)$ denotes the prior distribution of variable $Z$; 
$\mathbb{E}_{P(X,Y)}$ and $\mathbb{E}_{P(X)}$ denote data sampling that can be approximated by sample batches in optimization; 
hyperparameter $\beta$ controls the trade-off between reconstruction and disentanglement.
We prove in \Cref{th:equal to CAE} that CVAE is a special case of our DisGenIB without considering disentangled representation learning.
\begin{theorem}\label{th:equal to CAE}
For sample generation, DisGenIB is reduced to CVAE when the disentangled representation learning is ignored, \ie, the label-related information $A$ is absolutely provided by label variable $Y$ in $\mathcal{L}_{DisGenIB}$.
\begin{proof}
\renewcommand{\qedsymbol}{}
We use \Cref{prop:chain rule} and replace the label-related information $A$ with label variable $Y$ in objective $\mathcal{L}_{DisGenIB}$, arriving at 
{\small
\begin{align}
    \mathcal{L}_{DisGenIB}
    =& (2+\alpha)I(X;Y,Z)-(1+\alpha)I(X;Z)+\bar{C},
\end{align}}where the compression term degenerates to a constant $\bar{C}$.
To solve this optimization problem, we further estimate an upper bound of $I(X;Z)$ by
{\small
\begin{align}
\label{eq:equal to KL}
&I(X;Z) =\int P(X,Z)\log\frac{P(Z|X)}{P(Z)}\,dx \,dz\notag\\
&\leq \int P(X)P(Z|X)\log P(Z|X)\,dx \,dz-\int P(Z)\log R(Z)\,dz\notag\\
&=\mathbb{E}_{P(X)}\left[D_{KL}(P(Z|X)\| R(Z))\right],
\end{align}}where $R(Z)$ is a variational approximation to marginal distribution $P(Z)$. The inequality holds due to the non-negativity property of $D_{KL}(P(Z)\|R(Z))$.
Additionally, the upper bound of  $I(X;Y,Z)$ can be derived by
{\small\begin{equation}\label{eq:equal to E}
\begin{aligned}
I(X;Y,Z)\geq& \int P(X,Y,Z)\log Q_{\phi_2}(X|Y,Z)\,dx\,dy\,dz\\
=&\int P(X,Y)P(Z|X)\log Q_{\phi_2}(X|Y,Z)\,dx\,dy\,dz\\
=&\mathbb{E}_{P(X,Y)}\mathbb{E}_{P(Z|X)}\left[\log Q_{\phi_2}(X|Y,Z)\right].
\end{aligned}
\end{equation}}
According to \Cref{eq:equal to KL} and (\ref{eq:equal to E}), the proposed DisGenIB degenerates to the following optimization problem
\begin{align}
\label{eq:GenIB}
    \max_{Z}\  &\mathbb{E}_{P(X,Y)}\mathbb{E}_{P(Z|X)}\left[\log Q_{\phi_2}(X|Y,Z)\right]\notag\\
    &-\alpha^\prime \mathbb{E}_{P(X)}\left[D_{KL}(P(Z|X)\| R(Z))\right],
\end{align}
where $\alpha^\prime = \frac{1+\alpha}{2+\alpha}$. 
Note that the objective in \Cref{eq:GenIB} is consistent with CVAE.
The proof is completed.
\end{proof}
\end{theorem}

\subsection{Connection with attribute-based VAE (A-VAE)}
Given prior distribution $P(A)$, CVAE defined in \Cref{eq:cvae loss} turns to attribute-based VAE (A-VAE)  \cite{cheng2021disentangled,xu2022generating}, formulated as solving optimization problem
\begin{equation}
    \begin{aligned}
    \max_{Z} \mathcal{L}_{A-VAE}=&\mathbb{E}_{P(X,A)}\mathbb{E}_{P(Z |X)}\left[\log Q(X |Z, A)\right] \\
    -&\beta \mathbb{E}_{P(X)}\left[D_{K L}(P(Z |X) \| H(Z))\right].
    \end{aligned}
    \label{eq:beta-vae loss}
    \end{equation}
In this sense, the generation process is based on priors $A$ instead of label $Y$. A-VAE can therefore fully leverage priors to generate more faithful samples of unseen classes for FSL  \cite{xu2021variational}.
We prove in \Cref{th:equal to attribute VAE} that A-VAE is a special case of our DisGenIB with prior distribution $P(A)$.
\begin{theorem}
\label{th:equal to attribute VAE}
For sample generation, DisGenIB is reduced to A-VAE given prior distribution $P(A)$.
\begin{proof}
\renewcommand{\qedsymbol}{}
Given prior distribution $P(A)$, the objective of DisGenIB turns to be $\mathcal{L}_{DisGenIB}^{Prior}$ according to \Cref{prop:DisGenIB_prior}. Similar to the proof in \Cref{th:equal to CAE}, we get the consistency of two terms in $\mathcal{L}_{DisGenIB}^{Prior}$ and $\mathcal{L}_{A-VAE}$. The proof is completed.
    \end{proof}
\end{theorem}

Note that both CVAE and A-VAE fail to explicitly employ disentangled representation learning for generation. 
As a result, CVAE struggled in generating samples of unseen classes since their labels $Y$ are hardly available during training, while A-VAE cannot be implemented without priors. 
In contrast, our DisGenIB with disentangled representation learning relieves the distribution shift in generation, and further takes advantage of priors to facilitate the disentangled generation.

\section{FSL inference with DisGenIB}
\paragraph{Problem Definition.}
The training process of FSL is typically divided into two stages: pre-training and meta-test. In the pre-training phase, a base dataset is given as $\mathcal{D}_{base}=\{(\boldsymbol{x}_i,y_i):y_i\in\mathcal{C}_{base}, i=1,2,\cdots,D\}$ with substantial labeled samples of size $D$, where $\mathcal{C}_{base}$ refers to the base categories set; $\boldsymbol{x}_i\in\mathbb{R}^{d}$ denotes the $d$-dimensional feature of the $i$-th sample, associated with a label $y_i\in\mathcal{C}_{base}$. 
While in the meta-test phase of $N$-way $K$-shot FSL tasks, there are two sets sampled from novel category set $\mathcal{C}_{novel}$ ($\mathcal{C}_{base}\cap \mathcal{C}_{novel} = \emptyset$): support set and query set. 
The support set is defined as $\mathcal{S}=\{(\boldsymbol{x}_i,y_i):y_i\in\mathcal{C}_{novel},i=1,2,\cdots, N\times K\}$, where $N$ categories are sampled from the novel category set $\mathcal{C}_{novel}$ with $K$ labeled samples for each category. 
Correspondingly, the query set that is used to evaluate model's performance is denoted by $\mathcal{Q}=\{(\boldsymbol{x}_i,y_i):y_i\in\mathcal{C}_{novel},i=1,2,\cdots, N\times M\}$, where $M$ is the number of samples for each category. 
The main purpose of FSL is to learn an effective classifier for query set $\mathcal{Q}$ over the support set $\mathcal{S}$ as well as the base dataset $\mathcal{D}_{base}$.

\paragraph{Inference.}
We train DisGenIB on the base dataset $\mathcal{D}_{base}$ to learn a generative model for synthesizing an additional dataset $\mathcal{S}^\prime=\{(\boldsymbol{x}_i,y_i):y_i\in\mathcal{C}_{novel},i=1,2,\cdots\}$ with sufficient labeled samples of novel classes $\mathcal{C}_{novel}$.
Specifically, we utilize encoder $Q_{\theta_A}(A|X)$ to sample $A$ from the input,  or directly sample from prior distribution $P(A)$ when given priors. To enrich the diversity of generated samples, we utilize the encoder $Q_{\theta_Z}(Z|X)$ to obtain the sample-specific information from $\mathcal{S}\bigcup \mathcal{Q}$. Then we utilize the decoder $Q_{\phi_1}(X|A,Z)$ to generate samples of $\mathcal{S}^\prime$, whose labels are assigned by the corresponding label-related information. 

Motivated by  \cite{snell2017prototypical}, we learn a prototypical classifier with the help of DisGenIB and support samples, where prototypes are estimated as the feature centroid of each class.
In detail, we have two prototypes $\boldsymbol{p}$ and $\boldsymbol{p}^\prime$ that are estimated from $\mathcal{S}$ and $\mathcal{S}^\prime$, respectively. Similar to  \cite{zhang2021prototype}, we use Multivariate Gaussian Distribution (MGD) to model the mean and variance of these two prototypes, which are then fused to get the rectified prototype $\boldsymbol{\hat{p}}$ for evaluation.

\section{Experiments}
\subsection{Experimental Setup}
We evaluate DisGenIB on three FSL benchmarks, including miniImageNet, tieredImageNet and CIFAR-FS. Due to the page limit, the implementation and training details are given in supplementary.

\paragraph{Datasets.}
The miniImageNet and tieredImageNet, \ie, the subsets of ILSVRC-2012  \cite{deng2009imagenet}, are two standard benchmarks for the FSL task. More specifically, miniImageNet contains 100 classes with 600 images of size 84$\times$84 per class, while tieredImageNet includes 608 classes with around 1200 images of size 84$\times$84 per class. 
The CIFAR-FS benchmark used for FSL is a subset of CIFAR100   \cite{krizhevsky2009learning}. It contains 100 classes with 600 images of size 32$\times$32 per class. 
For fair comparison, we follow previous works  \cite{vinyals2016matching,ren2018meta,bertinetto2018meta} to split these datasets into training, validation and testing subsets, respectively.

\paragraph{Evaluation Metrics.}
In meta-testing, we run 5-way 1/5-shot classification tasks on 600 episodes randomly sampled from the test set, where 15 query images are randomly sampled per class for evaluation. 
For each experiment, we report the mean accuracy and the 95\% confidence interval.

\begin{table}[t]
\tabcolsep = 2 pt
    \begin{center}
    \begin{tabular}{lcccccc}
    \toprule 
    \multicolumn{1}{l}{\multirow{2}{*}{\textbf{Weight}}}
    & \multicolumn{1}{l}{\multirow{2}{*}{\textbf{BackBone}}}
    & \multicolumn{2}{c}{\textbf{CIFAR-FS}} 
    \\ 
    \cmidrule(l){3-4}
    \multicolumn{1}{c}{} & \multicolumn{1}{c}{}
    & \textbf{$1$-shot}   & \textbf{$5$-shot} \\
    \midrule
    \midrule
    MetaDT  \cite{zhang2022metadt} 
    & ResNet-12       & $79.03$            & $88.50$  \\
    RFS  \cite{tian2020rethinking}  
    & ResNet-12            & $71.51$         & $86.00$    \\
    PAL  \cite{ma2021partner}  
    & ResNet-12        & $77.10$            & $88.00$   \\
    CRF-GNN  \cite{tang2021mutual}  
    & CovNet-256      & $76.45$            & $88.42$ \\
    InvEq  \cite{rizve2021exploring}  
    &ResNet-12      & $76.83$            & $89.26$ \\
    CGCS  \cite{gao2021curvature}  
    &ResNet-12      & $71.66$            & $85.50$ \\
    MeTAL  \cite{baik2021meta}  
    &ResNet-12      & $67.97$            & $82.17$ \\
    SCL  \cite{ouali2021spatial}  
    &ResNet-12      & $76.50$            & $88.00$ \\
    RENet  \cite{kang2021relational}  
    &ResNet-12      & $74.51$            & $86.60$ \\
    SEGA  \cite{yang2022sega}  
    &ResNet-12      & $78.45$            & $86.00$ \\
    \midrule
    DisGenIB 
    & ResNet-12   & $85.19$   & $89.54$   \\
    DisGenIB + Prior
    & ResNet-12   & \textbf{86.84}   & \textbf{89.76}  \\
    \bottomrule
    \end{tabular}
    \end{center}
    \caption{Comparsion with SOTA methods on CIFAR-FS.}
    \label{tab:CIFAR-FS} 
    \end{table}

\begin{table}[t]
\begin{center}
\tabcolsep=10.6 pt	
\begin{tabular}{lcccccc}
\toprule 
\multicolumn{1}{l}{Proportion}
& \multicolumn{1}{l}{\textbf{mini}}
& \multicolumn{1}{l}{\textbf{tiered}}
& \multicolumn{1}{l}{\textbf{CIFAR-FS}} 
\\ 

\midrule
train  
& $28.15\%$       & $5.73\%$            & $31.25\%$  \\
test  
& $70.37\%$       & $12.09\%$            & $71.43\%$  \\
\bottomrule
\end{tabular}
\end{center}
\caption{Proportion of common attributes on different sets.}
\label{tab:attr proportion} 
\end{table}

\subsection{Comparison to State-of-the-Art}
In this section, we report the results of both our DisGenIB and current SOTA methods on FSL. We divide these SOTA methods into two categories, depending on whether they utilize priors.

\noindent\textbf{Evaluation on miniImageNet and CIFAR-FS.}
These two datasets have similar structures with randomly partitioned classes. As shown in \Cref{tab:mini_tiered_results} and \ref{tab:CIFAR-FS}, DisGenIB surpasses SOTA methods by more than 7\% in 1-shot tasks without priors. Compared to previous entangled generators AFHN and SVAE, the disentangled generation process of DisGenIB can maintain the discrimination of generated samples and thus greatly boosts the performance, \ie, 2\%-3\%. Other metric-based methods utilize various training strategies to learn robust features, \eg, cross-attention  \cite{kang2021relational}, feature reconstruction \cite{wertheimer2021few} and alignment \cite{xu2021learning}. Unfortunately, due to the entangled learning strategy, these methods inevitably introduce sample-specific noise into the learned features. In contrast, DisGenIB learns disentangled representations to mainly focus on the label-related information, which eliminates the impact of sample-specific information and thus outperforms these methods by 3\%-9\%.
Moreover, when priors are provided, our DisGenIB regards them as prior distribution to further facilitate the disentanglement. In contrast, other methods, \eg, TriNet and TRAML, only treat priors as extra features to rectify the classifier, which fail to fully use the priors. As a result, our DisGenIB exceeds these methods by more than 7\%.

\noindent\textbf{Evaluation on tieredImageNet.}
As shown in \Cref{tab:mini_tiered_results}, using priors or not, our DisGenIB is superior to SOTA methods by a large margin, \ie, 1\%-5\%. Nevertheless, compared to the other two datasets, the performance improvement over tieredImageNet is limited. We infer that this is mainly due to the partition of the dataset.
Specifically, tieredImageNet contains much more classes that need to be divided by high-level semantics, which is more challenging since the shared information among different partitions is limited. These issues undermine the generalization of the label-related information extraction $Q(A| X)$ and generation process $Q(X|A,Z)$. To quantitatively verify the above analysis, we report the shared information among different partitions of three datasets, which is evaluated by the proportion of the common attributes shared by train and test sets. As shown in \Cref{tab:attr proportion}, tieredImageNet has notably less common attributes between train and test sets than other two datasets. Thus, the performance over tieredImageNet is vulnerable to being diminished. We will explore this issue in future work.

\subsection{Ablation Study}
To further analyze the effect of each key component, we conduct substantial experiments on miniImageNet in 5-way 1-shot setting. Due to the page limit, more experiments can be found in the supplementary.

\noindent\textbf{Influence of generated samples.}
The quality of generated samples, \ie, discrimination and diversity, is the key to evaluating the generative models in FSL. To quantitatively study the efficacy of generated samples, we utilize them to rectify the prototype and linear classifiers. The former relies on the discrimination, while the latter depends on both discrimination and diversity of generated samples. To be more convincing, we also present the results of existing SOTA methods, \eg SRestoreNet   \cite{xue2020one}, BD-CSPN   \cite{liu2020prototype} and FSLKT   \cite{peng2019few}. \textbf{1)} We calculate the average cosine similarity between the rectified prototype $\boldsymbol{\hat{p}}$ and the true prototype over test set. As shown in \Cref{tab: proto_sim}, our DisGenIB surpasses these baselines by more than 0.15 over cosine similarity. \textbf{2)} We utilize the generated samples as extra samples to learn a linear classifier. As shown in \Cref{tab: cc_acc}, the classification accuracy of our DisGenIB exceeds the other three generative methods, \ie, AFHN  \cite{li2020adversarial}, IDeMe-Net  \cite{chen2019image} and VI-Net  \cite{luo2021few}, by more than 6\%. These results demonstrate the high discrimination and diversity of samples generated by our DisGenIB, which can effectively alleviate the data scarcity of FSL. 
\begin{table}
    \centering
        \subtable{
    \begin{minipage}[t]{0.47\linewidth}
    	\begin{center}
    		\begin{tabular}{lcccccc}
\toprule 
\multicolumn{1}{l}{Methods}
& \multicolumn{1}{l}{Sim}
\\ 

\midrule
Baseline  & 0.55      \\
SRestoreNet  & 0.79      \\
BD-CSPN   & 0.67      \\
FSLKT   & 0.68      \\
Ours  & \textbf{0.95}      \\
\bottomrule
\end{tabular}
    	\end{center}
    	\caption{Cosine similarity between the estimated and real prototypes. Here baseline is the original prototype on $\mathcal{S}$.}
    	\label{tab: proto_sim}
    \end{minipage}%
    }%
    \hfill
        \subtable{
    \begin{minipage}[t]{0.47\linewidth}
    	\begin{center}
    		    		\begin{tabular}{lcccccc}
\toprule 
\multicolumn{1}{l}{Methods}
& \multicolumn{1}{l}{Acc}
\\ 

\midrule
Baseline  & 52.73      \\
AFHN   & 62.38      \\
IDeMe-Net   & 59.14      \\
VI-Net   & 61.05      \\
Ours  & \textbf{68.36}      \\
\bottomrule
\end{tabular}
    	\end{center}
    	\caption{Classification accuracy of using generated samples with linear classifier. Here baseline is the  classifier learned on $\mathcal{S}$.}
    	\label{tab: cc_acc}
    \end{minipage}%
    }%
\end{table}

\begin{figure}[!t]
    \centering
\includegraphics[width=1\linewidth]{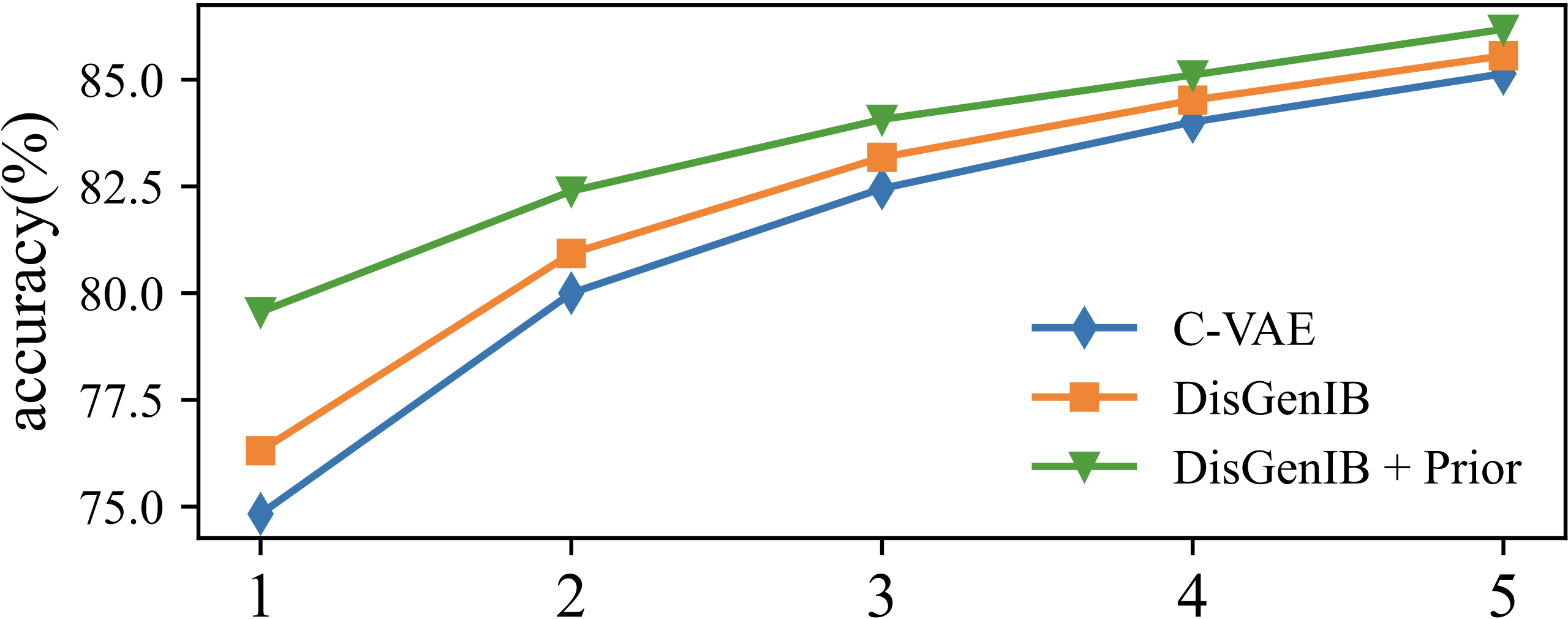}
\caption{Influence of disentanglement on different
number of support samples.}
\label{fig: disentagnlement}
\end{figure}

\begin{table}[t]
\tabcolsep = 4 pt
    \begin{center}
    \begin{tabular}{cc|cc|cc}
    \toprule 
    \multicolumn{2}{c|}{Regularizer} &
    \multicolumn{2}{c|}{miniImageNet} &
    \multicolumn{2}{c}{CIFAR-FS}
    \\
    \multicolumn{1}{c}{$I(Y;Z)$}
    & \multicolumn{1}{c|}{$\operatorname{IB}(A;X;Y)$}
    & \multicolumn{1}{c}{5w1s} 
    & \multicolumn{1}{c|}{5w5s} 
    & \multicolumn{1}{c}{5w1s} 
    & \multicolumn{1}{c}{5w5s} 
    \\ 
    
    \midrule
          &            & $71.52$  &$84.76$              & $83.13$  &$87.33$\\
     \checkmark       & & $73.95$  &$85.20$            & $84.07$  &$88.56$\\
            & \checkmark & $75.94$  &$85.31$            & $84.93$  &$88.87$\\
     \checkmark       & \checkmark & \textbf{76.31}  &\textbf{85.56}  & \textbf{85.19}  &\textbf{89.54}\\
    \bottomrule
    \end{tabular}
    \end{center}
    \caption{Comparative results of different regularizers on miniImageNet and CIFAR-FS. 5w1s and 5w5s stand for 5-way 1-shot and 5-way 5-shot, respectively.}
    \label{tab: ablation loss}
    \end{table}

\noindent\textbf{Influence of Regularizers.}
We propose a novel objective in \Cref{eq:DisGenIB} that consists of the conventional reconstruction term $I(X;A,Z)$, together with some regularization terms that control the disentanglement and compression of $A$ and $Z$, including $I(A;Y)-\beta I(X;A)$ (\ie, $\operatorname{IB}(A;X;Y)$) and $I(Y;Z)$.
As shown in \Cref{tab: ablation loss}, when removing both terms, the performance drops by around 3\% on average, mainly due to insufficient disentanglement on $A$ and $Z$. When merely using $\operatorname{IB}(A;X;Y)$, the performance significantly increases by around 2\% on average. In addition, due to the entanglement of label-related information, the improvement of using only $I(Y;Z)$ is limited since DisGenIB fails to maintain the discrimination of generated samples without $\operatorname{IB}(A;X;Y)$. The best performance is achieved when using both terms, demonstrating the effectiveness of the proposed regularizers in disentangling $A$ and $Z$.

\noindent\textbf{Influence of disentanglement.}
To study the effectiveness of the disentanglement, we report the results of our DisGenIB without disentanglement, \ie, CVAE. For comparison, we also report results of the original DisGenIB, as well as DisGenIB that uses priors to facilitate disentanglement.
As shown in \Cref{fig: disentagnlement}, with different numbers of support samples, the best disentangled ``DisGenIB + Priors" exceeds the other two methods by a large margin, while the worst disentangled CVAE drops drastically in performance. 
Thus, the performance is positively related to the degree of disentanglement, which is consistent with the study on regularizers.
These results demonstrate the necessity of disentanglement for maintaining the discrimination of generative models.

\section{Conclusion}
In this paper, we propose a novel IB-based disentangled generation framework to generate faithful unseen samples for FSL. Specifically, our DisGenIB solves disentangled representation learning and sample generation in a unified framework, together with effectively leveraging priors. Additionally, we theoretically prove that previous disentanglement and generative methods are special cases of our DisGenIB, which demonstrates the generality of DiGenIB. Benefiting from these, our DisGenIB achieves state-of-the-art performance on challenging FSL benchmarks. 
Our DisGenIB provides a novel and promising perspective on solving FSL tasks with interpretable information theoretic objectives.
\bibliography{aaai23}
\end{document}